\documentclass[runningheads]{llncs}
\usepackage[T1]{fontenc}
\usepackage{graphicx}
\begin{document}
\title{Deepfake Media Forensics: State of the Art and Challenges Ahead}
%
%
\author{Irene Amerini$^5$, Mauro Barni$^8$, Sebastiano Battiato$^1$, Paolo Bestagini$^6$, Giulia Boato$^9$, Tania Sari Bonaventura$^5$,
Vittoria Bruni$^5$, Roberto Caldelli$^2$, Francesco De Natale$^9$, Rocco De Nicola$^{10}$, Luca Guarnera$^1$, Sara Mandelli$^6$,
Gian Luca Marcialis$^4$, Marco Micheletto$^4$, Andrea Montibeller$^9$, Giulia Orrù$^4$, Alessandro Ortis$^1$, Pericle Perazzo$^3$,
Giovanni Puglisi$^4$, Davide Salvi$^6$, Stefano Tubaro$^6$, Claudia Melis Tonti$^5$, Massimo Villari$^7$, Domenico Vitulano$^5$
}
%
\authorrunning{I. Amerini et al.}
%
\institute{$^1$ University of Catania, $^2$ CNIT, Florence, and Universitas Mercatorum, $^3$ University of Pisa, $^4$ University of Cagliari, 
$^5$ Sapienza University of Rome, $^6$ Politecnico di Milano, $^7$ University of Messina,
$^8$ Università di Siena, 
$^9$ University of Trento, \\$^{10}$ Scuola IMT Alti Studi Lucca
}
\maketitle              
\begin{abstract}
AI-generated synthetic media, also called Deepfakes, have significantly influenced so many domains, from entertainment to cybersecurity. Generative Adversarial Networks (GANs) and Diffusion Models (DMs) are the main frameworks used to create Deepfakes, producing highly realistic yet fabricated content. While these technologies open up new creative possibilities, they also bring substantial ethical and security risks due to their potential misuse. The rise of such advanced media has led to the development of a cognitive bias known as Impostor Bias, where individuals doubt the authenticity of multimedia due to the awareness of AI's capabilities. As a result, Deepfake detection has become a vital area of research, focusing on identifying subtle inconsistencies and artifacts with machine learning techniques, especially Convolutional Neural Networks (CNNs). Research in forensic Deepfake technology encompasses five main areas: detection, attribution and recognition, passive authentication, detection in realistic scenarios, and active authentication. 
This paper reviews the primary algorithms that address these challenges, examining their advantages, limitations, and future prospects.

\end{abstract}
\section{Introduction}
The advent of Deepfakes, synthetic media generated by Artificial Intelligence (AI) that mimics real images, audio, and video, has significantly impacted various domains including entertainment, politics, and cybersecurity. Deepfakes leverage deep learning techniques, particularly GANs \cite{goodfellow2014generative} and DMs~\cite{ho2020denoising}, to create highly convincing but falsified representations of individuals. While these technologies offer creative opportunities, they also pose serious ethical and security challenges due to their potential for misuse.
The emergence of such advanced AI-generated media has led to the development of a cognitive bias known as the \textit{Impostor Bias}~\cite{CASU2024301795}, which refers to the tendency to doubt the veracity of multimedia elements due to the knowledge that they can be realistically generated by AI models.
Deepfake detection has become an essential field of research, aiming to develop methods to distinguish between real and artificially generated media. Techniques for Deepfake detection often involve analyzing inconsistencies and artifacts that are not easily perceptible to the human eye \cite{guarnera2020preliminary,corvi2023intriguing} but can be detected using proper detectors based on machine learning algorithms. These detection methods typically focus on both spatial and temporal anomalies in the data, utilizing Convolutional Neural Networks (CNNs) \cite{wang2020cnn,guarnera2023level} for enhanced accuracy. 
Starting with the Deepfake detection task, the scientific community has over the years taken on several other new challenges to study the nature of synthetic data in detail. We can therefore distinguish 5 main areas of research in the Forensic Deepfake domain, namely \textit{Deepfake Detection} (Section \ref{sec:MF}) \textit{Deepfake Attribution And Recognition} (Section \ref{sec:DAR}), \textit{Passive Deepfake Authentication Methods} (Section \ref{sec:PDAM}), \textit{Deepfakes Detection Method On Realistic Scenarios} (Section \ref{sec:DDMRS}), and \textit{Active Authentication} (Section \ref{sec:AA}).

In this context, authors of the proposed papers are involved in the FF4ALL initiative (FF4ALL - Detection of Deep Fake Media and Life-Long Media Authentication), 
which aims to develop theoretical and practical tools for detecting and combating media counterfeits or Deepfakes, tracing their origin and limiting their dissemination. 
In the following sections, a brief overview of the main algorithms that aim to address the above-mentioned challenges will be presented. 

\vspace{-.3cm}

\section{Deepfake Detection}
%
%
%
%



Deepfake technology poses significant challenges due to its potential for misuse, which can severely impact public well-being and trust. While current detection methods, primarily based on convolutional neural networks and deep learning paradigms, have shown promising results, they often struggle to generalize across the varied techniques employed in digital content manipulation. This issue primarily arises from the intricate interplay between textures and artifacts in Deepfake data, which traditional detection methods frequently overlook.
In this context, artifacts are unintentional distortions or irregularities that occur during the Deepfake generation process, including unusual pixel formations or edge anomalies. Conversely, textures refer to the inherent patterns and fine details present in authentic images, such as the natural appearance of skin and hair.
In fact, it is well established that synthetic manipulations typically disrupt the texture consistency of original images \cite{sun2020identifying} and often leave detectable traces in the form of artifacts in both spatial \cite{chai2020makes} and frequency domains \cite{durall2020watch}, particularly in specific facial regions \cite{tolosana2022Deepfakes}. Consequently, numerous studies focus their analysis on specific portions of face images to identify these inconsistencies.
One promising approach involves using both No-Reference (NR) and Full-Reference (FR) quality measures to detect subtle manipulations in video frames \cite{concas2024quality}. This method has significantly improved cross-manipulation generalization by focusing on areas susceptible to artifacts, such as the mouth and eyes, and analyzing the image quality degradation caused by Deepfake algorithms.
In addition to artifacts, texture analysis provides another robust basis for distinguishing between real and fake images. In some Deepfake technologies dedicated to face-swapping operations, the inner and outer faces have different identities, making texture inconsistencies particularly evident \cite{la20233d}.
However, focusing exclusively on either artifacts or textures in Deepfake detection can be limiting. While these approaches yield high accuracy in specific contexts, they often fail to adapt to new and evolving Deepfake techniques. To address this limitation, a novel framework called the Texture and Artifact Detector (TAD) has been proposed \cite{gao2024texture}. The TAD framework enhances Deepfake detection by leveraging both texture and artifact inconsistencies, thereby improving model generalization across various forgery scenarios through ensemble learning.
Unfortunately, the performance of these methods is often hindered by the challenges posed by highly compressed data. High compression ratios can obscure subtle manipulations, leading to a significant degradation in detection accuracy \cite{hu2021detecting}.
A promising solution involves leveraging a learnable adaptive high-frequency enhancement framework to enrich weak high-frequency details in compressed content, thereby enhancing the robustness of Deepfake detection under compression \cite{gao2024Deepfake}. Further details on compression impacts and related detection strategies will be discussed in subsequent sections.


\label{sec:MF}

\section{Deepfake Attribution and Recognition}
\label{sec:DAR}
%
%
%
%





\subsection{Deepfake Fingerprint and Attribution}

Deepfake attribution, often referred to as Deepfake Model Recognition~\cite{guarnera2022exploitation,pontorno2024exploitation}, encompasses methodologies capable of identifying the specific model used to generate synthetic data. This process includes attempts to estimate the unique model weights~\cite{asnani2023reverse} of the architecture instance responsible for creating the Deepfake. SOTA techniques are highly effective in detecting Deepfake content generated by widely-used GANs~\cite{guarnera2020Deepfake,guarnera2020fighting,giudice2021fighting}  and DMs~\cite{guarnera2024mastering,pontorno2024deepfeaturex}. These techniques can even specialize to recognize the specific architectures, and, in more details, the specific model used in the creation procedure. Then, a more advanced challenge in this domain is identifying the exact model instance, characterized by a unique set of weights and parameters, within a given architecture: Guarnera et al.~\cite{guarnera2022exploitation} demonstrated that using a simple ResNET-18~\cite{he2016deep} engine combined with a metric learning approach~\cite{liu2012metric}, excellent results can be achieved in identifying the specific model used for creating synthetic data from 100 different instances of StyleGAN2-ADA~\cite{NEURIPS2020_8d30aa96}. 
A robust model recognition solution would enable the attribution of an image to a specific model owner, which is crucial for intellectual property rights~\cite{leotta2023not}. To establish the ownership or authenticity of an image generated by a particular model within a specific architecture, new strategies and appropriate metrics are required~\cite{huang2023can}.
In the context of forensic investigations involving Deepfake images, videos, or audio, state-of-the-art Deepfake detectors and architecture classifiers can be likened to the task of identifying camera models in traditional forensic analysis. 
Deepfake model recognition aims to trace the origin of a Deepfake to a specific model instance within an architecture. This parallel underscores the necessity of developing advanced techniques for Deepfake model attribution to ensure authenticity in digital media.

\section{Passive Deepfake Authentication Methods}
\label{sec:PDAM}
%
%
%
%


In the modern era, where video calls have become a cornerstone of global communication, the importance of authenticating audio and video streams cannot be overstated. The advent of Deepfake technology poses a significant challenge to the integrity of digital communication.
Traditional Deepfake detection methods may fall short as they often focus on either audio or video data in isolation. However, Deepfakes may involve sophisticated manipulations of both audio and video streams, making them harder to detect with monomodal methods.

This highlights the need for a multimodal approach that simultaneously analyzes both audio and visual data \cite{salvi2023robust}. By correlating information from these two channels, we can significantly improve the accuracy of Deepfake detection. This approach takes advantage of the fact that inconsistencies are often more noticeable when multiple modalities of data are considered together. For instance,~\cite{hosler2021Deepfakes} leverages the incongruity between emotional cues portrayed by audio and visual modalities. In~\cite{agarwal2023watch}, the authenticity of a speaker is verified by detecting anomalous correspondences between his/her facial movements and what he/she says. Moreover, the results of~\cite{khalid2021evaluation} show that an ensemble of audio and visual baselines outperforms monomodal counterparts.

Given that audio-visual authentication methods may exploit monomodal detectors in a synergistic fashion, another step towards better performance is to enhance monomodal audio or visual detectors separately through the use of advanced techniques. Concerning the visual component, it is possible to leverage fusion of multiple detectors trained on purpose to capture orthogonal traces \cite{mandelli2022detecting}. Concerning audio, it is possible to exploit modern solutions such as transformers \cite{singhyadav2024mdrt}, as well as investigating the use of semantic traces \cite{attorresi2022combining}.
Despite the great effort of the multimedia forensics community, a series of challenges remains. Concerning multimodal solutions, the need for audio-video Deepfake datasets is becoming more than an urgent necessity. Indeed, most of the effort has been put towards monomodal datasets creation. Moreover, given the trend of large language models, it could be interesting to try using the same logic for audio visual reasoning. Concerning monomodal solutions, explainability has definitely not been reach yet, which still proves a problem in case of court of laws.



\section{Deepfakes Detection Method on Realistic Scenarios}
%
%
%
%


\vspace{-.2cm}
\subsection{Deepfake Detection of image-videos in the wild } 
\label{subsec.wild}
\vspace{-.2cm}
In recent years, there has been a growing interest in the study of techniques for the detection of Deepfake and AI-generated media \cite{corvittofaiella2023,maianosp2024}. Consequently, numerous solutions have been proposed to address the problems posed by the increasing spread of fake multimedia content. However, most of these solutions perform well only in controlled settings, such as laboratory experiments, but fail to provide reliable results in real-life conditions typical of practical applications.
Deep Learning (DL) models can effectively detect Deepfake media and identify their source. However, despite promising results, DL-based methods face several significant challenges, particularly in real-world applications where controlled laboratory conditions are absent. Firstly, DL models require vast amounts of labeled data for training, which is often difficult to obtain in real-life scenarios. Additionally, these models must handle unforeseen situations that were not accounted for during training, a common issue in multimedia forensics applied outside the lab. For DL models to be effective in the ongoing battle between forensic analysts and counterfeiters, it is crucial to address the risk of overfitting to training data, which can lead to failures in new, unexpected situations. While it is possible to train highly accurate detectors, these methods struggle to generalize to new generative techniques due to data drift \cite{Paleyes2022}. Detectors perform well on the techniques they were trained on but often fail with content from new generative models. Another major obstacle is the black-box nature of DL techniques, making it difficult to interpret analysis results and understand decision-making processes. This lack of transparency hampers the practical application of DL in scenarios where accountability is essential. To address these issues, researchers address their attention on several strategies, including the use of one-class classifiers trained only on pristine images \cite{khalid2020,rs16050781}, developing classifiers with rejection options to opt out when encountering unfamiliar inputs not well-represented in the training set \cite{OOD_abstrain,siamese_based_dfd} and adopting methods capable of generalization though features fusion \cite{LEPORONI202499} together with multimodal approaches that combine audio and video streams  \cite{MongelliMA24,MajidICIAP2023}. 
Even considering these efforts, the practical application of automatic detectors has been minimal. Deploying these tools in commercial or mass verification systems presents numerous challenges beyond generalizing from a few known benchmarks \cite{Rossler20191,corvi2023intriguing,Li20231339,Zi20202382}. One significant challenge is the need to continuously train these models on new generative/Deepfake techniques in a continual learning fashion \cite{tassone2024continuous}.
Continual learning, also known as lifelong or incremental learning, is an ongoing approach to maintaining good model performance on evolving tasks without experiencing \emph{catastrophic forgetting} \cite{French1999128}. This approach is well-suited to recognize content generated by new techniques and continuously adjusting models to account for data drift observed during inference versus training.
A promising direction is the creation of an end-to-end Deepfake detection system that supports continuous integration and continuous delivery/deployment (CI/CD) with the design of a Machine Learning Model Operations (\emph{MLOps} \cite{Paleyes2022,semola2022continual}) pipeline, enabling the end-to-end development of continuously trained and monitored intelligent detectors with a minimal set of components.

\subsection{Deepfake and Social Media} 
An extremely challenging ``real-world" case scenario involves the detection of Deepfake multimedia shared on social networks \cite{10007988,8397040,pasquini2021media}. In fact, to cope with bandwidth and storage limitations \cite{10007988}, social networks apply severe data compression and resizing. However, such processing, while reducing the overall multimedia size, also reduces the presence of forensic features used for the discrimination of real vs. fake multimedia \cite{10007988,Verdoliva2020910,maier2024reliable}.
A first study on GAN images \cite{8397040} shared on Twitter demonstrates the adversarial effects of social network compression on Deepfake detectors. Specifically, while the visual quality of the shared images was untouched, the presence of forensic traces and patterns was reduced. The effects of social network sharing were then extensively observed and studied in \cite{10007988}.
In \cite{pasquini2021media}, the authors explore the challenges and advancements in the field of media forensics as applied to social network. The study addresses the increasing concerns regarding the authenticity and reliability of digital media shared on social platforms, focusing on challenges affecting source attribution algorithms \cite{lukavs2006detecting} as well as multimedia verification \cite{Verdoliva2020910}. For the latter, additional effort was devoted to studying whether the multimedia content is consistent with its descriptive text. The paper \cite{pasquini2021media} discusses the emerging challenges in the field, such as the rise of Deepfakes and the use of bots for spreading disinformation. The authors highlight the need for advanced forensic tools to keep up with these sophisticated methods of media manipulation.
In \cite{10007988}, the authors composed a large and diverse dataset counting 80k fake images generated with StyleGAN models and 70k real images collected from several state-of-the-art datasets \cite{karras2019style}. In addition to this, while revealing insightful details on the entity and severity of social network compression applied by Twitter, Facebook and Telegram, the authors shown how their dataset can be used to finetuning new detectors, preserving their accuracy on social network compressed images while without experiencing ``\textit{catastrophic forgetting loss}" \cite{French1999128}.
Interestingly, \cite{baracchi2024towards} showed that while social networks degrade the presence of forensic artifacts used by real vs. fake detectors, they introduce other traces that can be exploited to reconstruct the life cycle of multimedia and determine on which social networks it has been shared. While the life cycle of multimedia does not provide specific information on its nature (i.e., real or fake), it can be fundamental in recovering the version of that multimedia closer to the original, unshared one. This, in turn, allows for more accurate real vs. fake detection.
Finally, preliminary studies have been conducted also on videos shared on social netowrks \cite{marcon2021detection}, showing similar effects to those on images. One of these works \cite{marcon2021detection,9408664} studies the effects of social network compression on FaceForensics \cite{marcon2021detection} videos shared on Facebook and Youtube. The study provides a results in line with what observed on images \cite{marcon2021detection} and a new dataset of shared videos to be used to finetune real vs. fake detectors.
Nevertheless, while new works on Deepfake detection on social networks are available, the continuous update of social network compression algorithms makes the arms race increasingly challenging. As a consequence, this require additional effort to develop new architectures and updated datasets of social network shared images.

\subsection{Detection of Deepfake Images and Videos in Adversarial Setting}

An additional problem affecting virtually all the Deepfake forensic techniques developed so far is that such techniques are thought to operate in a benign setting, that is, by neglecting the possible efforts made by an adversary to mislead the forensic analysis. Yet, recent researches \cite{SzeAdv,Good14} have shown how easy is to generate adversarial contents capable of deceiving image and video processing techniques based on DL, when the adversary is informed about the details of the tools employed by the analyst. Some works \cite{Pap17,Bar19} have also studied the transferability of adversarial examples to networks different than those targeted by the attack, opening the way to the development of powerful attacks even when the attacker is unaware, or only partially aware, of the techniques used by the analyst.
Even worse, often it is not even necessary that the adversary applies sophisticated attacks relying on the full or partial knowledge of the to-be-attacked system. By relying on the lack of robustness and the generalization capabilities of the forensic tools already outlined in Section \ref{subsec.wild}, the adversary may simply process the deepkfake content in such way to prevent a correct forensic analysis, or at least degrade its performance up to a point to make it unusable. Some examples of this kind of attacks, often referred to as {\em laundering attacks}, include the application of moderate to strong lossy compression, geometric processing of images and videos, noise addition, histogram stretching and many others. 

Understanding and ensuring the security of Deepfake forensic tools is a crucial problem, if such tools have to be used under the intrinsically adversarial conditions typical of multimedia forensics applications.
For this reason, several efforts have been made to defend against adversarial attacks \cite{Bount23,Liang22}, both in the realm of computer vision applications and multimedia forensics. Still, no general effective solutions have been found yet \cite{ACW18}. Among the solutions developed so far, adversarial training \cite{Madry17} has received some consensus and has proven to at least mitigate the effectiveness of adversarial attacks in computer vision applications. As argued in \cite{Tsi18}, adversarial training forces DL models to focus on robust, possibly semantic, features, which are inherently more difficult to attack. Whether such a beneficial effect of adversarial training also applies to Deepfake forensic applications is still an open problem. It is no clear, in fact, if in multimedia forensics the equivalent of semantic computer vision features exist or not.

With regard to laundering attacks, the solutions proposed so far are similar to those already discussed in Section \ref{subsec.wild}, given the, ultimately, the effectiveness of laundering attacks can be drastically reduced by improving the robustness and generalization capabilities of the forensic tools. A common approach to do so, involves the use of data augmentation techniques that enrich the training set with processed samples, thus improving the robustness of the forensic tools against the processing operators included in the data augmentation procedure. Yet, accounting for {\em all} possible kinds of processing during training is clearly unfeasible. Among the solutions proposed so far, we mention the possibility of identifying a kind of {\em worst possible laundering attack} and include it in the training procedure. Examples of such an approach are described in \cite{BNT17,Purn24}. Despite all the efforts made, even for laundering attacks, a definitive solution has not been devised yet, thus adding yet another point to the {\em to-do} list of multimedia forensic researchers.

\label{sec:DDMRS}

\section{Active Authentication}
%
%
%
%


\subsection{Active deepefake detection}
Passive Deepfake detection techniques \cite{Caldelli2024,Caldelli_wacv2024} work a posteriori, after that the forged content has been generated, distributed and possibly processed, on the contrary, active methods work in a preemptive way, pre-processing the media in such a way to ease the subsequent analysis. This is the case, for instance, of Deepfake detection methods based on DNN watermarking, whereby the content generated by DNNs is watermarked in such a way to ease the distinction between genuine and fake media, and the attribution of the fake content to the network which generated it. An alternative possibility is to modify the computational imaging chain characterizing modern acquisition devices, to insert within the generated content a unique fingerprint to be used later on for authentication purposes. This means a change of paradigm that needs to be properly explored.  Active authentication techniques represent a valid, and more reliable, alternative (or complement) to passive authentication, whenever the operating conditions allow their use.

Watermarking has recently been proposed as a means to protect the IPR of DNNs \cite{Tondi_active22}. By tying a watermark to a DNN model, in fact, it would be possible to prove the ownership of the model or trace its illegal use. On this basis, DNN watermarking can be also used to link AI-generated contents \cite{Yu_21}, like Deepfakes, to the model which generated them, thus providing an easy and convincing  way to distinguish between synthetic and natural contents. Such a goal is achieved by requiring that all the contents generated by a network contain a predefined watermark (a kind of synthetic fingerprint) that can be used later on to distinguish the synthetic images (or videos) generated by a trained model from real ones. This marks a drastic paradigm change with respect to current solutions based on multimedia forensics, since authentication is now achieved with the active help of the party which trained the media-generation network. Though some solutions have appeared in this direction, putting this idea at work requires that considerable advances are made particularly in terms of watermark robustness against image, video and network manipulations, security against adversarial attacks, payload and also imperceptibility. Succeeding into designing a new class of robust and secure solutions, based on active approaches, for Deepfake detection/attribution surely represents an open challenge and an interesting opportunity for scientific research in the field of multimedia forensics.

\subsection{Efficient Media Origin Authentication}

Customary deep-fake detection methods, both passive and active, are subject to false positives and false negatives, whose rate highly depends on the employed method and the goodness of the training data. False positives are due to various factors, such as the complexity of the content, the quality of the training data, or the intrinsic limitations of the detection algorithm itself. False negatives could happen if the deep-fake is very well made, or in general if the detection method fails to recognize certain patterns or features that indicate a deep-fake.
On the other hand, cryptographic signatures are ``almost perfect'' from this point of view, in the sense that false negatives (i.e., authentic signatures which are not recognized so) are zero and false positives (i.e., fake signatures that are taken as authentic) are considered computationally infeasible to forge. This suggests that cryptographic signatures could be used fruitfully to detect deep-fakes (or better, the absence of deep-fakes) with perfect precision.
In this direction, the work-in-progress standard JPEG Trust \cite{temmermans2023towards} by the Joint Photographic Experts Group (JPEG) aims to establish trust in digital media by addressing aspects of authenticity, provenance, and integrity. JPEG Trust will provide a framework for establishing trust in media through secure annotation of media assets throughout their life cycle, using cryptography as a key component. Cryptographically signing media allows deep-fakes to be repudiated by the interested person, since the signature on them will be absent or invalid. Such an anti-fake signature should allow ``good'' manipulation of original file (at least cropping), disallow ``bad'' manipulation, but also be space efficient, to save bandwidth on web servers once the media file is disseminated. Unfortunately, customary signature schemes like ECDSA do not have these properties. 

To address this challenge, a solution could rely on novel aggregatable signatures, such as the Boneh-Lynn-Shacham (BLS) signature \cite{boneh2001short,ietfdraftBls}, which has been successfully used in blockchain technologies like Ethereum 2.0 to optimize storage\footnote{https://github.com/ethereum/consensus-specs/blob/dev/specs/phase0/beacon-chain.md\#bls-signatures}. The BLS signature scheme makes use of a novel form of cryptography called pairing-based cryptography, which allows for a plethora of new functionalities like attribute-based encryption\cite{goyal2006attribute,rasori2018abecities,sicari2021attribute}. Aggregatable signatures could be employed in JPEG (possibly within the JPEG Trust standard itself) in such a way to permit benign alterations of the image like cropping while preventing malicious tampering, without increasing too much the bandwidth occupation on web servers.

\subsection{Trusted Remote Media Processing on Cloud and Edge Computing Systems}

In the emerging Smart Cities context, systems based on IoT (Internet of Things) play an important role to allow citizens to interact with the environment and to benefit from advanced services, such as video surveillance, intelligent traffic lightning, and air quality sensing. From a technological point of view, using sensors and actuators to automate services is strategic, but managing, configuring, and optimizing the digital infrastructures to adapt their behavior to the specific needs of the context is a big challenge, both in terms of system design and security. Deepefake media detection in these scenarios represent a challenge due to the nature of possible manipulations of future citizens day life, hence a more holistic approach should be considered where the media production occur, since at the Edge Computing Systems \cite{9912968}.  
In just 20 years, with the objective to increase system response and reduce communication latency, computation moved from mainframes and computing rooms towards Cloud Computing, Fog Computing, and lastly, Edge Computing. A Federated Cloud-Edge infrastructure is considered, where different administrative domains are in place and where Machine Learning software artifacts, in the assertion of Federated Learning even at the Edge, help to distribute intelligence in this scenario.

Multimedia acquisition devices based on IoT generate an unprecedented amount of data, with the need of developing Cloud-based video big data analytics frameworks. A distributed approach in video recording and elaboration systems, such as video surveillance systems based on IP cameras, is highly recommended to overcome the maximum storage or throughput limitation of Network Video Recorders installed on single machines. 
To perform such a variety of tasks, and to be able to modify a device’s behavior on-demand, the Function as a Service (FaaS) computational paradigm is generally adopted. FaaS allows to define several minimal applications and to run one or more instances of these on the same device at the same time.
FaaS framework relies on two configuration approaches: a local configuration file, generally YAML, or a secure remote server. However, both come with limitations: a local file configuration requires direct access to the device, physically or through a secure connection, to modify it.
Alternatively, a remote server can store and send updated configuration files, but it might be vulnerable to well-known cyberattacks suck as Man-in-the-middle (MITM) or Distributed Denial of Service (DDoS), making it unusable and unreliable.
To overcome such limitations, it is possible to benefit from three technologies that have been increasingly recognized to be able to address information access problems and system trustiness in different application domains: Federated Learning \cite{10413933}, Blockchain and IPFS (InterPlanetary File System) starting at the Edge:
\begin{itemize}
    \item Federated Learning is a decentralized approach to training Machine Learning Models. In traditional Machine Learning, data is centralized in the Cloud, where a single model is trained on the entire dataset. Federated Learning, on the other hand, allows for training Machine Learning Models across multiple decentralized devices or servers that hold local data samples without exchanging them. Moreover, Federated Learning at the Edge refers to the application of Federated Learning techniques on Edge devices, such as IoT devices, or Edge Servers. This approach combines the benefits of Federated Learning, which ensures Data Privacy and reduces communication costs, with the advantages of Edge Computing Systems, which enables data processing and model training to occur closer to where the data is generated, hence, fake Media might not exit from the Edge. 
    \item The use of Blockchain, supported by the flexibility and robustness of Smart Contracts, allows the combining of the well-known FaaS paradigm with the intrinsic features of data non-repudiation and immutability, replacing the service configuration with a Smart Contract, guaranteeing protection against distributed cyber-attacks \cite{9631479}. 
    \item IPFS is a distributed system for storing and accessing files. Since the block size of the Blockchain does not allow storing files, these can be uploaded to this special file storage, which produces a unique hash value to be used as a key to access its content \cite{10181167}.

\end{itemize}

\label{sec:AA}

\section{Conclusion}
\label{sec:conclusion}
In this paper, we have conducted a comprehensive review of the state-of-the-art techniques and challenges in Deepfake media forensics. Our exploration covered the core areas of Deepfake detection, attribution and recognition, passive authentication, detection in realistic scenarios, and active authentication. Each of these areas addresses specific facets of the Deepfake phenomenon, from the identification of synthetic media and tracing their origins to ensuring the robustness of detection systems in real-world environments and embedding verifiable information within media for instant authentication.
Future work will focus on conducting a more in-depth analysis of practical countermeasures and gaining deeper insights into real-world applications (e.g. highly compressed data). 

\begin{credits}
\subsubsection{\ackname} 
This study has been partially supported by SERICS (PE00000014) under the MUR National Recovery and Resilience Plan funded by the European Union - NextGenerationEU.

\subsubsection{\discintname}
The authors have no competing interests. 
\end{credits}
%
%
%
%

\bibliographystyle{splncs04}
\bibliography{main}

\end{document}